\documentclass[10pt,twocolumn,letterpaper]{article}

\usepackage{comment,hhline}
\usepackage{float}
\usepackage{cvpr}
\usepackage{times}
\usepackage{epsfig}
\usepackage{graphicx}
\usepackage{amsmath}
\usepackage{amssymb}
\usepackage{gensymb}
\usepackage{enumitem}
\usepackage{multirow}

\usepackage[pagebackref=true,breaklinks=true,letterpaper=true,colorlinks,bookmarks=false]{hyperref}

\cvprfinalcopy 

\def\cvprPaperID{882} 

\ifcvprfinal\fi

\begin{document}

\title{Camera Calibration from Dynamic Silhouettes Using Motion Barcodes}

\author{Gil Ben-Artzi ~~~~~~~ Yoni Kasten ~~~~~~~ Shmuel Peleg ~~~~~~~~~ Michael Werman\\
School of Computer Science and Engineering\\
The Hebrew University of Jerusalem, Israel
}

\maketitle
\thispagestyle{empty}

\begin{abstract}

Computing the epipolar geometry between cameras with very different viewpoints is often problematic as matching points are hard to find. In these cases, it has been proposed to use information from dynamic objects in the scene for suggesting point and line correspondences.

%

We propose a speed up of about two orders of magnitude, as well as an increase in robustness and accuracy, to methods computing epipolar geometry from dynamic silhouettes. This improvement is based on a new temporal signature: motion barcode for lines.
Motion barcode is a binary temporal sequence for lines, indicating for each frame the existence of at least one foreground pixel on that line. The motion barcodes of two corresponding epipolar lines are very similar, so the search for corresponding epipolar lines can be limited only to lines having similar barcodes. The use of motion barcodes leads to increased speed, accuracy, and robustness in computing the epipolar geometry.

\end{abstract}

\section{Introduction}

\begin{figure}[t]
\begin{center}
 \includegraphics[width=1\linewidth]{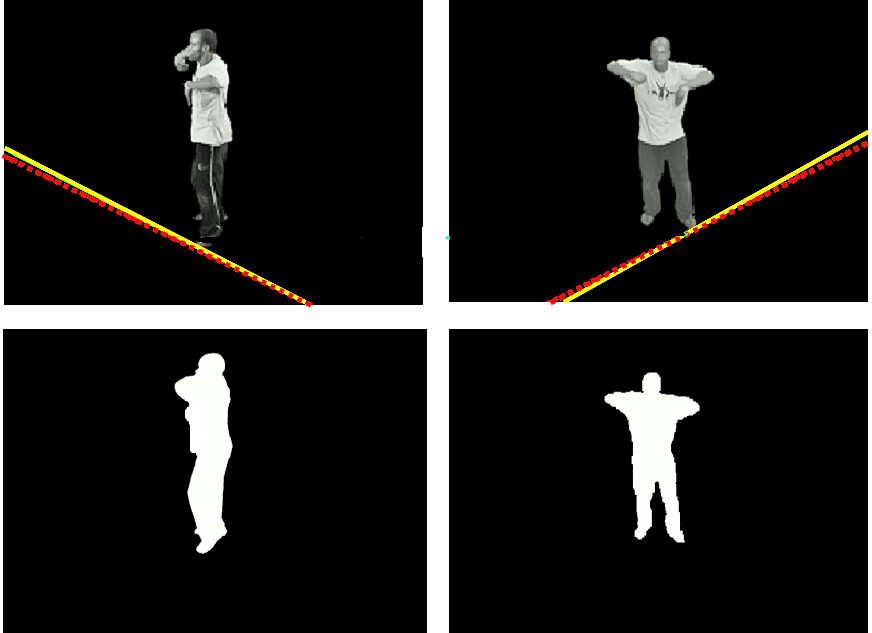}\\
\end{center}
\caption{When two cameras have very different viewpoints as in this example, appearance can not be used for calibration. Instead, calibration is possible from matching pairs of epipolar lines that can be extracted efficiently from moving silhouettes. The yellow lines are the epipolar lines proposed by our method, while the red lines are the ground truth epipolar lines. The corresponding silhouettes are displayed at the bottom.}
\label{fig:intro}
\end{figure}

Calibration of multi-camera systems is normally computed by finding corresponding feature points between images taken by these cameras. When not enough feature points can be found, e.g. when the camera viewpoints vary greatly, the epipolar geometry can be computed from silhouettes of moving objects that are visible in videos captured by the two cameras. The  silhouettes at one time instance are used to suggest matching epipolar lines which are used to propose a fundamental matrix that is verified over all frames.

The best methods for computing the fundamental matrix use tangents to the dynamic silhouette as candidates for epipolar lines \cite{sinha2010camera}. Our approach presents a speedup of about two orders of magnitude for these methods, and significantly improves accuracy and robustness. This speedup is obtained by requiring candidates for matching epipolar lines to share a temporal signature (motion barcode).

Motion barcodes were first introduced for points in \cite{benevent}. The motion barcode of a line is a binary temporal sequence, indicating for each frame the existence of at least one foreground pixel on that line. We show that correlation between the motion barcodes of corresponding epipolar lines is high. By testing as possible matches only pairs of lines whose motion barcode correlation is high, a speedup by about two orders of magnitude is obtained.
Figure~\ref{fig:intro} shows matching epipolar lines extracted using our approach. Following \cite{sinha2010camera}, we use a RANSAC approach to test possible matching pairs of epipolar lines, and compute the epipolar geometry.


This paper is organized as follows. Section~\ref{sec:prior} describes relevant prior work.
Section~\ref{sec:background} introduces the theoretical background. Section~\ref{sec:signatures} presents the motion barcodes of lines. Section~\ref{sec:epipolarLines} shows how to match epipolar lines  based on dynamic silhouettes. Section~\ref{sec:optimize} presents an iterative computation of the fundamental matrix based on the motion barcode. Section~\ref{sec:results} shows our results on both synthetic and real sequences.

\subsection{Prior Work}
\label{sec:prior}

Extracting geometrical information from the motion of silhouettes include shape-from-silhouettes \cite{cheung2003shape,forbes2006shape,aganj2007spatio} and camera calibration \cite{hernandez2007silhouette,sinha2010camera,boyer2006using,ramanathan2000silhouette,zhang2009self}. In shape-from silhouettes, the goal is to recover the visual hull \cite{laurentini1994visual,miller2006exact} of the object. If the cameras are calibrated, this task is relatively clear as each individual viewing cone \cite{franco2003exact} can be backprojected and the visual hull is the intersection of these cones. 

The case of uncalibrated cameras has also been investigated, where the goal is to recover epipolar geometry. The first step is to establish correspondences between special points on the silhouettes boundaries, called \emph{frontier points} \cite{cipolla2000visual}, across the different views.  These points are images of object points that are tangent to an epipolar plane. Given corresponding frontier points, spatial constraints resulting from matching epipolar tangents \cite{schmid1998geometry} are used to recover the epipolar geometry. 

Matching corresponding frontier points and silhouette tangents can be found using robust estimation procedures such as RANSAC \cite{fischler1981random}. Matching frontier points, or directions of four epipolar tangent lines, are initially guessed.  Furukawa et al. \cite{furukawa2006robust}, assuming orthographic projection, match frontier points using RANSAC. They used the distances between parallel tangent lines on the silhouettes as a geometric measure  for matching.
Given the epipoles and the accurate tangent envelope of the silhouettes, frontier points can be easily matched using two outermost epipolar tangents.  This property was deployed by Wong and Cipolla \cite{wong2001structure} for turntable motion. The most relevant previous work is Sinha and Pollefeys \cite{sinha2010camera}, addressing projective projection. They propose a RANSAC based search of possible epipoles, where a proposed epipole in each of the two corresponding images is generated from the intersection of two lines randomly selected from the tangent envelope.  
 
Calibration without explicitly matching tangent epipolar lines has also been considered. \cite{boyer2006using} used constraints based on the back projection of silhouettes boundaries in multiple views. \cite{yamazoe2006multiple} jointly optimized the 3D position of  frontier points and the camera parameters in a bundle adjustment. However, both methods require a good initialization of silhouette boundaries and camera parameters. Hernandez \cite{hernandez2007silhouette} also proposed constraints based on the back projection of silhouettes for maximizing silhouette coherence, but his method is limited to turntable motion.	 

Binary temporal signatures of pixels which are based on the motion of the objects in the scene have been previously introduced. Ermis et al. \cite{ermis2010activity} deploy such features to find accurate correspondence between pixels across distributed cameras with the assumption of a distant, almost planar, scene. Drouin et al. \cite{drouin2010camera}  matched 2D points between a video projector and a digital camera. They require a planar surface and the same ordering of pixels across views. The closest work related to the motion barcode is  \cite{pundik2010video} where  the {\it line signal}, the sum of intensity values of pixels on epipolar lines after background subtraction, was defined. This was used for video  synchronization. The line signal depends on motion and color and was used assuming known calibration. Ben-Artzi et al. \cite{benevent} introduce a method to match events across different views even in the case of significant parallax and occlusions. However, that approach can not be applied to match pixels for camera calibration, as its localization is very inaccurate, as explained in Section~\ref{sec:signatures}. Using a temporal histogram as a temporal feature of a pixel is introduced in \cite{Hoshen-Temporal}, but it is effective only for objects that are static for substantial periods. 


\section{Theoretical Background} 
\label{sec:background}

\begin{figure}[t]
\begin{center}
 \includegraphics[width=1\linewidth]{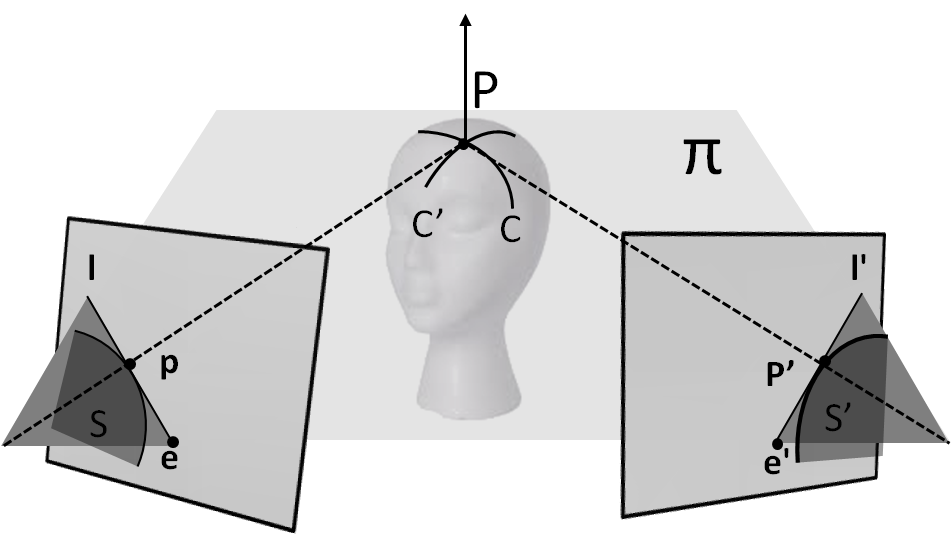}
\end{center}
   \caption{The geometry of two views with a tangent epipolar plane and a frontier point. $\Pi$ is the tangent epipolar plane and $l,l'$ are tangent epipolar lines in $\Pi$. The epipolar plane is tangent to the object at the frontier points $P$. }
\label{fig:frontier}
\end{figure}

The geometric relation between corresponding silhouettes across views is based on frontier points and epipolar tangencies \cite{mendonca2001epipolar,cipolla1995motion,lazebnik2001computing}.

The geometry of two views containing silhouettes is presented in Fig.~\ref{fig:frontier}. 
For the rest of the paper let {\em candidate points} be image points that are on the boundary of the silhouette as well as on the boundary of its convex hull.
$C$ and $C'$ are the contours of the object in 3D. These contours project to  silhouette boundaries $S$ and $S'$. The two contours intersect in the frontier point $P$. The projections of P onto the two views are the candidate points, p and p'. The 3 points $P,p,p'$ span a \emph{tangent} epipolar plane $\Pi$ between the two views. The points $p$ and $p'$ must lie on the corresponding tangent lines $l$ and $l'$ and the point $P$ is the location where the tangent epipolar plane $\Pi$ is tangent to the surface, $e,e'$ are the epipoles. The frontier points are the only true corresponding points between the boundaries $S$ and $S'$. If we have accurate tangents to the silhouettes and the location of the epipole is known, then the epipolar tangent lines give the corresponding points $p$ and $p'$. This idea was traditionally used as a \emph{spatial} cost function between corresponding tangential epipolar lines and points. See for example \cite{furukawa2006robust,sinha2010camera}.

Finding frontier points without the epipole locations is difficult \cite{porrill1991curve}.  For a video sequence, when the location of at least two frontier points are known, their tangent lines are epipolar lines. They can be used to calculate the epipole and the location of the other frontier points in all the other frames.  Alternatively, if the epipoles are known we can use the tangent lines to the silhouettes to locate the frontier points. It follows that either the frontier points or the epipoles are needed in order to extract the epipolar lines.

Here we introduce a different approach which does not require  prior knowledge of either frontier points or epipoles in order to extract matching epipolar lines.  We directly compute epipolar line correspondences using the motion observed simultaneously by them. 

\section{Motion Barcode: Temporal Signature of Lines}
\label{sec:signatures}

\begin{figure}[t]
\begin{center}
 \includegraphics[width=1\linewidth,height=0.4\linewidth]{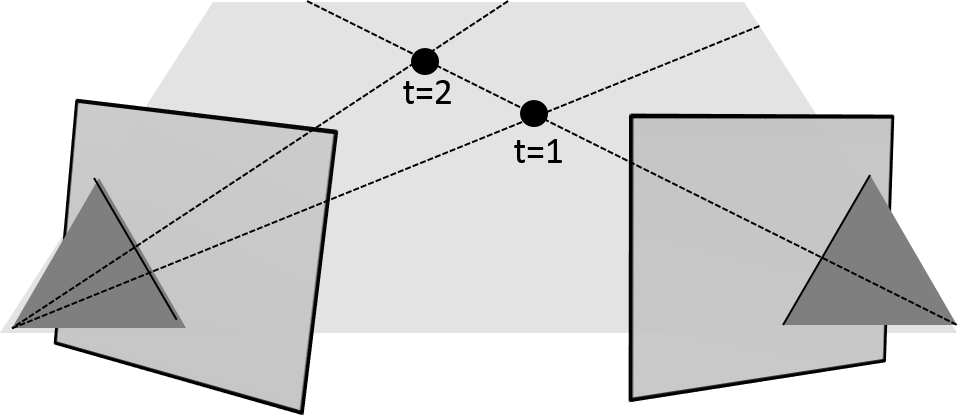}
\end{center}
   \caption{In dynamic scenes, the geometrical relation between pixels is characterized only up to corresponding epipolar lines. Each pixel in one video can correspond to different pixels at different times in the other video. For stationary cameras, the different pixels in the second view will always reside on the epipolar line. }
\label{fig:p2p}
\end{figure}

\begin{figure}[t]
\begin{center}
 \includegraphics[width=1\linewidth,height=0.6\linewidth]{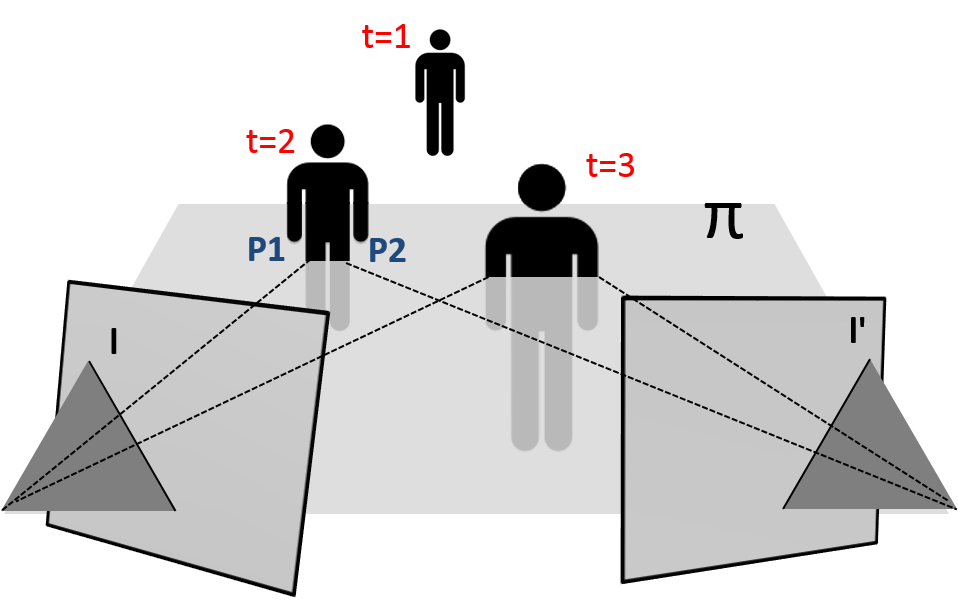}
\end{center}
   \caption{The motion barcodes of two corresponding epipolar lines, $l$ and $l'$, in a video of a moving person at three time instances. If a point on an epipolar line is a projection of a foreground point at time $t$, then there exists a point on its corresponding epipolar line which is also a projection of a foreground point at the same time.  In the figure, at time $t=2,3$ the two corresponding epipolar lines contain a point from a silhouette. This can be a different 3D point due to viewpoint differences, e.g.  $P_1,P_2$. The motion barcode of both epipolar lines in this figure, $b_l$ and $b_{l'}$, is [0,1,1].}
\label{fig:temporalEpipolar}
\end{figure}

Given two frames captured at the same time from different viewpoints, two corresponding pixels view a single 3D point. However, in a dynamic scene, a  pixel in one view is bound to correspond to different pixels of the other view at different times, located on the corresponding epipolar line. 

Fig.~\ref{fig:p2p} illustrates  a typical case. At time $t=1$, a single pixel in the right view corresponds to some pixel in the left view. At time $t=2$, due to the motion of the object in scene, the same pixel corresponds to a different pixel in the other view. For video sequences captured by stationary cameras the corresponding pixels will always reside on corresponding epipolar lines.    

It follows that if an epipolar line contains \emph{at least} one silhouette pixel at time $t$, then its corresponding epipolar line should contain such a pixel at the \emph{same time}. This is illustrated in Fig.~\ref{fig:temporalEpipolar}. For time $t=2,3$ there are points from objects that project onto the corresponding epipolar line. If a point on line $l$ is part of a silhouette, this point or another silhouette point occluding it, will be seen on line $l'$. Alternatively, the silhouette could be blocked by a background object or be out of the frame. 

The motion barcode of a line $l$, $b_l(t)$, indicates for each line $l$ in frame $t$ the existence of at least one foreground pixel on that line. $b_l(t)=1$ if the line intersects a silhouette, and $b_l(t)=0$ otherwise.
The motion barcodes of two corresponding epipolar lines is very similar. Differences occur only in cases of occlusions.


The \emph{temporal} similarity between two lines $l$ and $l'$ is  defined as the correlation between their motion barcodes;

\begin{equation}
d_t(l,l^\prime )=corr(b_l(t),b_{l^\prime}(t))
\label{Eq:temporal1}
\end{equation}

\section{Epipolar Geometry by Matching Lines}
\label{sec:epipolarLines}

\begin{figure}[t] 
\begin{center}
\includegraphics[width=1\linewidth,height=0.35\linewidth]{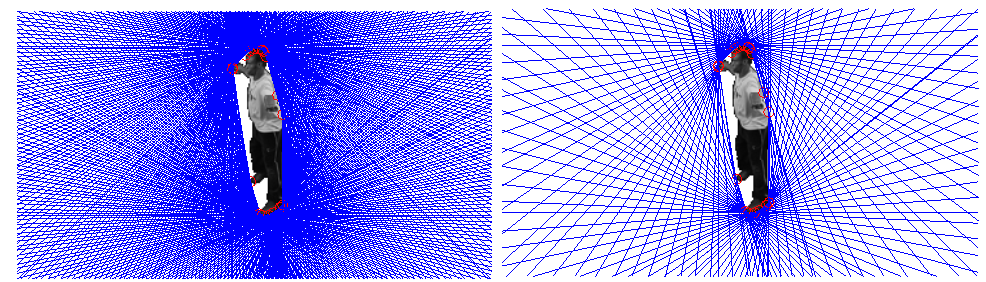}
 (a)~~~~~~~~~~~~~~~~~~~~~~~~~~~~~(b)
\end{center}
\caption{Uniform sampling of lines from the tangent envelope. (a) Lines sampled every $1\degree$, (b) Lines sampled every $4\degree$  } 
\label{fig:epipolarTangentEnvelope}
\end{figure}

The epipolar geometry can be computed from 3 pairs of corresponding epipolar lines \cite{hartleyMVG}. The search space for matching epipolar lines across views is very large if we consider all possible pairs of lines. This search can be reduced to fewer lines by using only {\em candidate lines}, lines tangent to the {\em tangent envelope} of the silhouette. The tangent envelope includes points that are on the silhouette boundary as well as on the boundary of its convex hull. We follow the work of \cite{sinha2010camera}, checking for possible correspondence only lines on the tangent envelope. Using only candidate lines is justified as the projection of the frontier point is on the tangent envelope. 


\begin{figure}[t] 
\begin{center}
 \includegraphics[width=1\linewidth,height=0.4\linewidth]{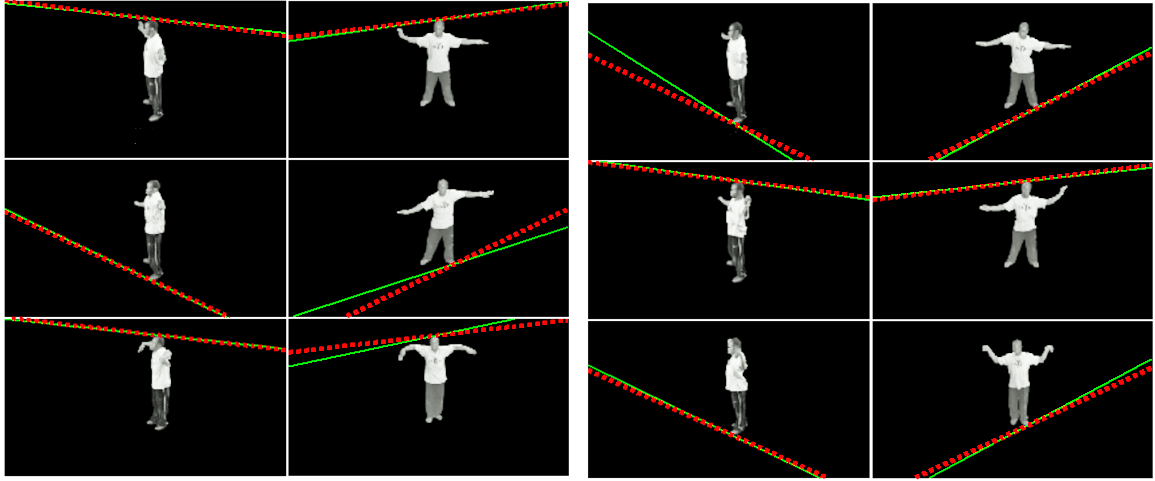}
\end{center}
   \caption{Finding corresponding epipolar lines by their motion barcode. Every pair of frames contributes one possible match. The dashed lines are the true epipolar lines and the green lines are the candidate pairs having highest barcode correlation.}
\label{fig:matchingLines}
\end{figure}

We select several corresponding pairs of frames from the video sequences, so that the pairs will be sufficiently different from each other. For each pair of frames, we sample $K$ candidate lines from the tangent envelope of its silhouettes. We compute the correlation between the motion barcodes for all pairs of candidate lines from the two corresponding images. This results in $K^2$ correlations per each pair of frames. From every pair of frames we select the single pair of epipolar lines with the highest barcode correlation. Fig.~\ref{fig:epipolarTangentEnvelope} shows all candidate lines, and Fig.~\ref{fig:matchingLines} shows the pairs of candidate lines having highest barcode correlation. We compute the epipolar geometry from three matching pairs based on \cite{hartleyMVG}. The computation is by RANSAC similarly to \cite{sinha2010camera}. The fundamental matrix is then fully optimized as described in Section~\ref{sec:optimize}.


The matching is carried out in two phases. An offline phase where the motion barcodes of the tangent lines are computed.  In the online phase, the actual matching is carried out by computing the correlation between motion barcodes of pairs of lines, and computing the epipolar geometry using RANSAC. The overall efficiency depends mainly on accuracy of candidate matches as it effects the number of required iterations in the RANSAC phase. Details are in Section~\ref{sec:results}.

\section{Temporal Optimization of the Fundamental Matrix}

\label{sec:optimize}
Existing optimization techniques for computing the fundamental matrix are based on minimizing a spatial cost function without taking into account the temporal dimension.  We present  a technique  based on both spatial and  temporal cost functions (Eq.~\ref{Eq:temporal1}). 

We  assume a set of corresponding points, presumably the projection of frontier points $\{(x_i,x_i')\}_{i=1}^M$, a set of corresponding epipolar lines $\{(l_i,l_i')\}_{i=1}^N$ and an initial estimation of the fundamental matrix $F$. The optimization is   iterative. In the first step we  optimize the point correspondences based on the lines, using  the geometric reprojection error \cite{hartleyMVG} as the spatial cost function. In the second step we  optimize the epipolar line correspondences based on the given points, using the temporal cost function (Eq.~\ref{Eq:temporal1}).  We optimize the directions of the epipolar lines for each pair of corresponding points.  Based on the lines matched in this step, we estimate epipoles and an epipolar line homography. We then evaluate a set of corresponding points and obtain an estimation of the fundamental matrix.  The process is described in the following:
\begin{itemize}
\item Step one:
\begin{enumerate}[leftmargin=*]
\item Minimize reprojection error based on $\{(x_i,x_i')\}_{i=1}^M$. 
$$ \sum_i d(x_i,\hat{x}_i)^2+d(x_i',\hat{x}_i')^2~~~s.t.~~~\hat{x_i} \hat{F} \hat{x_i}'=0 $$
This minimization is by the Levenberg-Marquardt procedure and gives a new set of points and fundamental matrix. 
\item set $l_i=\hat{F} \hat{x_i}', l_i'=\hat{F}^T \hat{x_i}$.
\end{enumerate}
\item Step two:
\begin{enumerate}[leftmargin=*]
\item For each pair of lines, minimize  
\begin{equation}
C_l(\hat{l},\hat{l}')= d_s(l,\hat{l})+d_s(l',\hat{l}')-d_t(\hat{l},\hat{l}')
\label{optimizeLineDirection}
\end{equation}
$d_s$ measures the angular deviation between lines, and $d_t$ is the barcode correlation (Eq.~\ref{Eq:temporal1}). $d_s$ ensures the lines are within an angle difference of no more than $\Theta$. The choice of $\Theta$ will be discussed next. $\hat{l},\hat{l}'$ are sampled uniformly from  $[-\Theta,\Theta]$ around $l,l'$.  We take the maximal match and if we have more than one maximum, the one with the minimal angle difference is selected.
\item Estimate new epipoles $e,e'$ and epipolar line homography from 
$\{\hat{l}_i,\hat{l}_i'\}$.
\item Set $\{x_i,x_i'\}$  by projecting onto the nearest $l,l'$. Estimate $F$ from epipoles and lines homography.
\end{enumerate}
\end{itemize}

The process  terminates when the deviation of the estimated epipoles is small enough or a maximum number of iterations is exceeded.

The choice of the angular tolerance $\Theta$ defines the region where we look for the newly estimated lines. It  depends on the epipolar envelope and the required probability for locating the line \cite{unger2010new}. Direct modeling of epipolar envelope is difficult and therefore it is empirically evaluated, see \cite{csurka1997characterizing,xu1996epipolar,hartleyMVG}. In our implementation we set $\Theta$ to $0.2\degree$ which results in an accurate estimation. This reflects our assumption that the distortion is low. The specific choice can be adjusted according to the needs.    

\begin{figure}[tb]
\begin{center}
    \includegraphics[width=0.25\linewidth]{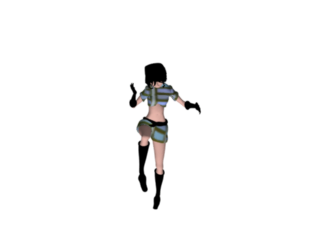} 
   \includegraphics[width=0.25\linewidth]{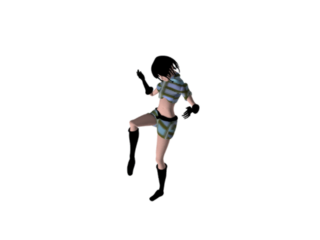}
   \includegraphics[width=0.235\linewidth]{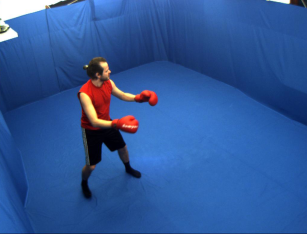}
   \includegraphics[width=0.235\linewidth]{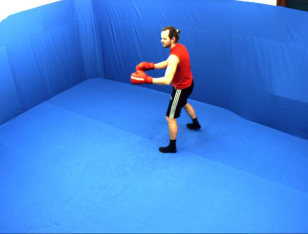}\\
   (a)~~~~~~~~~~~~~~~~~~~~~~~~~~~~~~~~~~~~~~~~~~~~~~(b)\\
   ~~~~~~\includegraphics[width=0.07\linewidth]{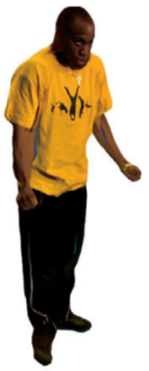}~~~~~~~~~~~~~~
   \includegraphics[width=0.08\linewidth]{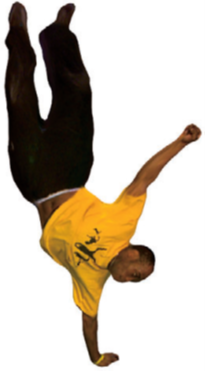}~~~~~~~~~~~~
   \includegraphics[width=0.235\linewidth]{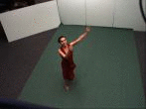}
   \includegraphics[width=0.235\linewidth]{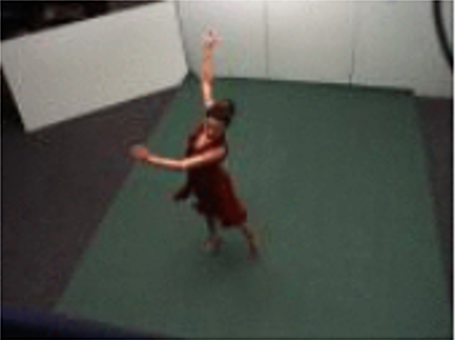}\\
   (c)~~~~~~~~~~~~~~~~~~~~~~~~~~~~~~~~~~~~~~~~~~~~~~(d)
  
\end{center}
\caption{The datasets used in the experiments. (a)  The synthetic Kung-Fu girl dataset. (b) The Boxer dataset. (c) The Street Dancer dataset (d) The Dancing Girl dataset. }
\label{fig:datasets}
\end{figure}

\begin{table}[b]
\begin{center}
\begin{tabular}{ |l|l|l|l|}
\hline
	  Dataset & Type & Camera Pairs & Frames \\ \hline         
      KungFu Girl   & Synthetic &  300 &  200   \\
      Boxer         & Real &       6 &   778   \\
      Street Dancer & Real &       15 &   250  \\
      Dancing Girl  & Real &       28 &   200  \\
\hline
\end{tabular}
\end{center}
\caption{ Dataset properties
\label{table:datasetProperties}}
\end{table}

\section{Experiments}
\label{sec:results}

Our approach was validated on synthetic and real sequences. We compared our method with the state of the art method \cite{sinha2010camera}, where the fundamental matrix is computed by RANSAC-based sampling of epipolar lines. The evaluation was done with the following datasets: the Kung-Fu girl \cite{kungfu}, Boxer \cite{Boxer}, Street Dancer \cite{StreetDancer} and Dancing Girl \cite{DancingGirl}. Fig.~\ref{fig:datasets} shows images from the datasets and Table ~\ref{table:datasetProperties} gives the details.  

We compared the accuracy and efficiency of the two methods. The accuracy of the fundamental matrix in all experiments is measured by the symmetric epipolar distance (error) \cite{hartleyMVG} using ground truth matching points. The symmetric epipolar distance is the distance between each point and the epipolar line corresponding to the other point. The acquisition of the ground truth points is discussed in Subsection~\ref{subsec:impl}.

The efficiency of the methods is evaluated as follows. In both methods the fundamental matrix is computed using RANSAC sampling of epipolar lines. In each iteration, the symmetric epipolar distance of each hypothesis is evaluated.  Every 1000  RANSAC iterations the best hypothesis is selected and optimized using the non-linear Levenberg-Marquardt (LM) optimization procedure as in \cite{sinha2010camera}. The efficiency is  measured by the number of non-linear optimization procedures required to reach a given accuracy (error). The less non-linear optimization procedures the more efficient the method is. A detailed description is in Subsection~\ref{subsec:impl}.

\begin{table}[tb]
\scriptsize
\begin{center}
\begin{tabular}{ |l|l||r|r|r|r|r|r|}
\hline
\multicolumn{8}{|c|}{No. of Non-Linear Optimizations Needed to Reach a Desired Accuracy} \\ \hline
\multicolumn{2}{|c||}{Sym Epipolar Distance } & 1.5  & 1 & 0.8 & 0.5 & 0.4 & 0.3 \\ 
\hhline{|==#=|=|=|=|=|=|}
\multirow{2}{*}{Kung-Fu} 
 & Ours     &1&2&4&23&71&302 \\
 & Sinha    &19&65&134&822&1989&8659\\
\hline
    
\multirow{2}{*}{Street Dancer} 
& Ours   & 3&7&20&255&616&1233   \\
& Sinha  & 37&159&340&1871&7485&-   \\     
\hline

\multirow{2}{*}{Dancing Girl} 
& Ours &2&4&9&129&918&13776  \\     
& Sinha &36&149&388&13972&-&- \\     
\hline

\multirow{2}{*}{Boxer} 
& Ours   &2&5&12&111&996&- \\     
& Sinha   &333&2994&2994&-&-&- \\     
\hline

\end{tabular}
 \end{center}
\caption{  
The expected number of non-linear optimizations required to reach a given accuracy of the fundamental matrix. Accuracy is measured using symmetric epipolar distance with respect to ground-truth points. The best hypothesis is selected every 1000 RANSAC iterations, and is further optimized using non-linear (LM) method. In each dataset, the number of optimizations is averaged over all cameras pairs.  Empty cells indicate that the required accuracy was not attained.
\label{table:iterations}}
\end{table}

\begin{figure}[tb]
\begin{center}
   \includegraphics[width=1\linewidth]{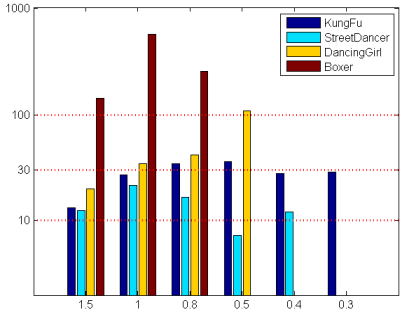} \\
 \end{center}
\caption{     
The ratio between our method and Sinha \cite{sinha2010camera} of the number of non-linear optimization procedures required to reach a given fundamental matrix accuracy. The horizontal axis is the accuracy in terms of the desired symmetric epipolar distance of ground truth points.
\label{fig:ratio}}
\end{figure}

There is a difference in the error used during RANSAC and the error we use for final evaluation. During RANSAC, the quality of an hypothesis is evaluated based on inliers, as ground truth is unknown. This error is usually lower from the error of ground truth points. We used ground-truth points for a non biased evaluation.

{\bf Efficiency }  The expected number of non-linear LM optimization procedures required to reach a fundamental matrix having a better accuracy than a predefined level is shown in Table~\ref{table:iterations}. For each pair of cameras, we executed 500K RANSAC iterations resulting in 500K hypotheses. Every 1000 RANSAC iterations the best hypothesis is selected and optimized non-linearly. The accuracy of the optimized fundamental matrix, in terms of the  symmetric epipolar distance of ground truth points, is recorded. The accuracy values after all non-linear optimization procedures from all camera pairs in the dataset form our samples. For example, in the Kung-Fu dataset we executed 500K$\times$300 RANSAC iterations, performed 500$\times$300 LM optimizations, and collected 150,000 samples. We  build the cumulative distribution function (cdf) of the error from all camera pairs. Given the cdf the expected number of samples is extracted.  
It can be seen that our method quickly converged to sub-pixel accuracy. Fig.~\ref{fig:ratio} shows the ratio between the required number of non-linear optimization procedures in our method and Sinha\cite{sinha2010camera}. The horizontal dashed lines are in ratios of 10, 30 and 100. For accuracy of 0.8 pixel, the median of the ratios between the required number of non-linear optimization procedures is 38, and for accuracy of 1.5 pixel the median of the ratios is 17. 

\begin{table}[bt]
\scriptsize
\begin{center}
\begin{tabular}{ |l|l||r|r|r|r|r|r|}
\hline
\multicolumn{8}{|c|}{Symmetric Epipolar Distance} \\ \hline
\multicolumn{2}{|c||}{RANSAC Hypotheses}& 1K & 2K  & 5K & 10K & 20K & 100K \\ 
\hhline{|=|=#=|=|=|=|=|=|}
\multirow{2}{*}{Kung-Fu} 
 & ours  & 1.11&0.85&0.64&0.54&0.47&0.35 \\
 & Sinha &  4.9&3.41&2.12&1.63&1.29&0.78  \\
\hline
    
\multirow{2}{*}{Street Dancer} 
& ours     &1.93&1.29&0.97&0.85&0.75&0.59   \\
& Sinha   & 4.31&3.35&2.4&1.96&1.59&1.01  \\     
\hline

\multirow{2}{*}{Dancing Girl} 
& Ours & 1.4&1.09&0.83&0.72&0.63&0.49   \\     
& Sinha &6.28&4.57&2.96&2.15&1.6&1   \\     
\hline

\multirow{2}{*}{Boxer} 
& Ours &1.63&1.46&0.85&0.74&0.65&0.48 \\     
& Sinha &7.06&5.82&4.02&3.37&2.8&1.86 \\     
\hline

\end{tabular}
 \end{center}
\caption{  
Accuracy reached by each method for a fixed number of RANSAC samples. Accuracy is after a non-linear optimization phase, measured with respect to ground-truth points. 
\label{table:accuracy}}
\end{table}

\begin{table}[tb]
 \footnotesize
\begin{center}
\begin{tabular}{  c |  c |  c|  }    
  & Ours  &  Sinha  \\ \hline
Kung-Fu          & 0.26 & 0.51               \\
Street Dancer    & 0.36  & 0.62    \\ 
Dancing Girl    & 0.41  & 0.72   \\ 
Boxer            & 0.39    &1.33           \\ 
	\end{tabular}
\end{center}
\caption{The best accuracy reached by each method on all camera pairs in each dataset after 500K RANSAC hypotheses. The accuracy is the median over all camera pairs of the best symmetric epipolar distance reached after the non-linear optimization phase. 
\label{table:maxAccuracy}}
\end{table}

\begin{figure}[tb]
\begin{center}
   \includegraphics[width=0.49\linewidth]{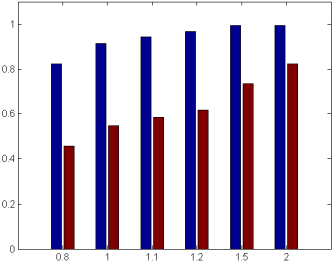} 
   \includegraphics[width=0.49\linewidth]{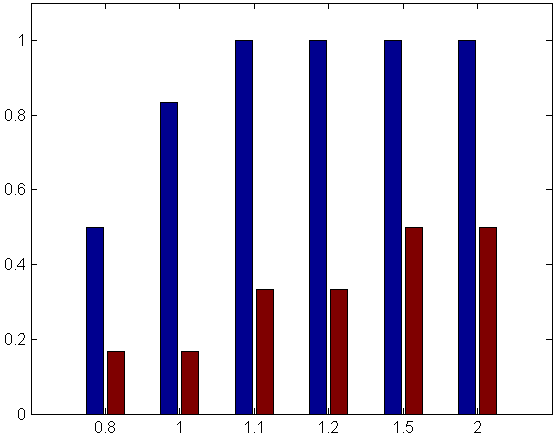} \\
   (a)~~~~~~~~~~~~~~~~~~~~~~~~~~~~~~~~~~(b)\\
   \includegraphics[width=0.49\linewidth]{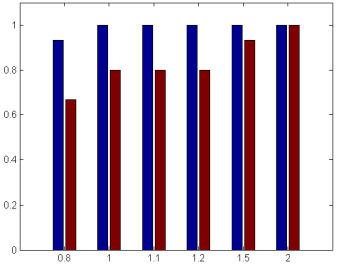} 
   \includegraphics[width=0.49\linewidth]{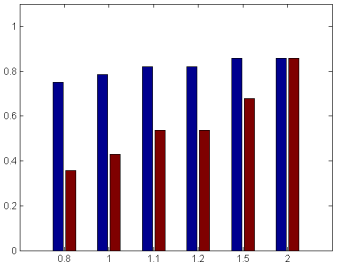} \\
   (c)~~~~~~~~~~~~~~~~~~~~~~~~~~~~~~~~~~(d)\\
 \end{center}
\caption{The fraction of camera pairs whose fundamental matrices reached a given symmetric epipolar distance. The accuracy is evaluated over 500K RANSAC iterations. The x-axis is the given accuracy. The y-axis is the fraction of camera pairs that reached this accuracy.  The blue bars are our method and the red bars are Sinha's method. (a) The Kung-Fu dataset. (b) Boxer dataset. (c) Street Dancer. (d) Dancing Girl. 
\label{fig:numCameras}}
\end{figure}

{\bf Accuracy} We evaluated the best accuracy (minimal error) reached for a given number of RANSAC generated hypotheses. For each pair of cameras in the dataset, we generated 500K RANSAC hypotheses by each method. We subdivided the hypotheses into equal sized groups. From each group we selected the best hypothesis (lowest symmetric epipolar distance) with respect to the ground truth points. We then applied non-linear optimization and measured the accuracy of the resulting fundamental matrix. The  accuracy is the median over all optimized fundamental matrices.  For example, in the Kung-Fu dataset we have 150,000 hypotheses. For evaluation of the highest accuracy reached by 5K hypotheses, we divided them into 30 equal size groups, optimized the best hypothesis from each group and evaluated the median over the symmetric epipolar distances. Table~\ref{table:accuracy} shows the results. It can be seen that for the Kung-Fu dataset, our method requires approximately 2K RANSAC iterations followed with  1 non-linear optimization procedure to reach an accuracy of 0.85. Using our approach, in less than 5K RANSAC iterations all datasets reached sub-pixel accuracy. Table~\ref{table:maxAccuracy} shows the best median accuracy reached by each method.
As expected, the synthetic dataset has best accuracy, 0.26, while the worst accuracy, 0.41, was in the Dancing Girl dataset which has many errors in the silhouettes.   
        
We also evaluated the fraction of the number of camera pairs whose fundamental matrices reached a given accuracy using all the samples, after the non-linear phase. The results are shown in Fig.~\ref{fig:numCameras}.  For the Kung-Fu dataset, for 298 out of 300 camera pairs the accuracy reached $1.5$, including pairs where the cameras are facing each other. This is discussed in the next subsection. On average, the number of camera pairs where a given accuracy was reached using our method is by a factor of 1.8 higher than the number of cameras with same accuracy using Sinha's method. The average is calculated over all camera pairs over all datasets. 

\begin{figure}[tb]
\begin{center}
\includegraphics[width=0.5\linewidth]{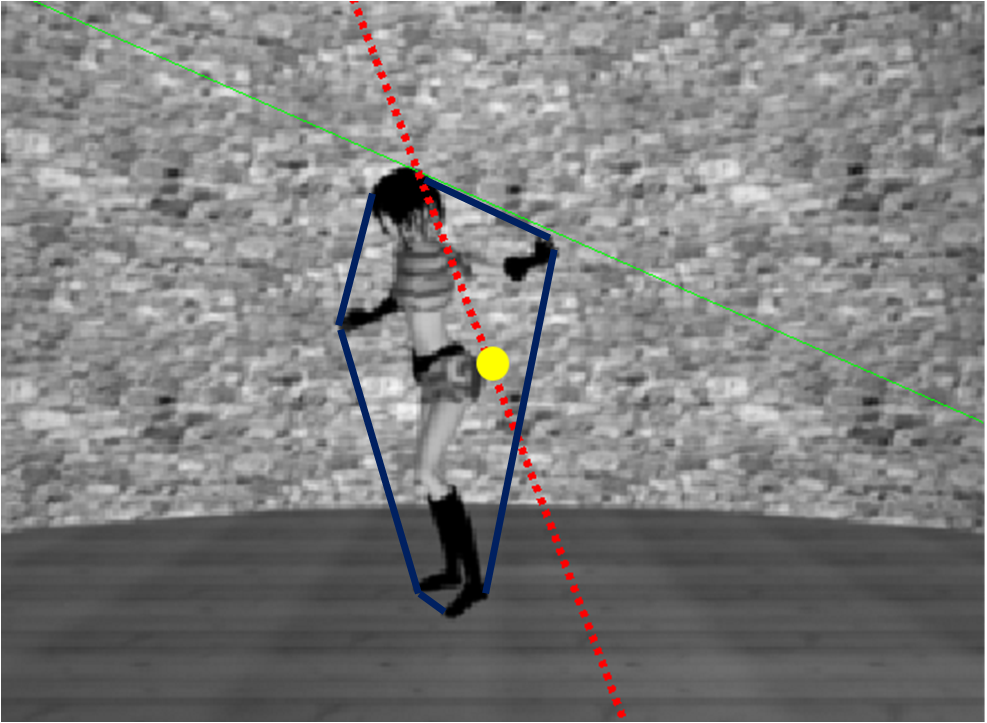} 
  \end{center}
\caption{ When the epipole is at the center of the image, e.g. when two cameras are facing one another, it may not be possible to find epipolar lines. In this case the epipole is often inside the convex hall. In this example the convex hull is marked in blue, and the yellow point is the epipole. The red line is a ground truth epipolar line. The green line is an hypothesized epipolar line.}
\label{fig:epipoleInside}
\end{figure}

\subsection{Frames Lacking Frontier Points}

Frames that lack frontier points are problematic for most tangent based methods.
This happens when the epipoles are inside the convex hull of the dynamic objects, a common case when the two cameras face each other. An example is illustrated in Fig.~\ref{fig:epipoleInside}. Using our method, even when the pairs of cameras are facing each other, the fundamental matrix can still be recovered. This is possible as the object is moving, and there are often a few frames where the epipole if outside the convex hall. These few frames are enough for the calibration, due to the accuracy of the selected candidates for epipolar lines.
For example, it can be seen in Fig.~\ref{fig:numCameras} that in the Kung-Fu dataset, for accuracy of $1.5$, our method fails for only two camera pairs, whereas Sinha's method fails on 78 camera pairs.

\subsection{Ground-Truth Error vs. Inlier Error}

The symmetric epipolar distance \cite{hartleyMVG} is a quality measure for fundamental matrices, and is defined over a set of pairs of corresponding points across two images. 

In ordinary computations of the fundamental matrix, when no ground truth data is known, the symmetric epipolar distance is calculated based on \emph{hypothesized} inlier points. Since some inliers are often wrong correspondences, there is a significant difference between the error computed on inlier points and the error computed on ground truth points (when available). As we have access to the ground truth in our datasets, we used the ground truth points to measure the symmetric epipolar distance and evaluate our experiments.  

\subsection{Implementation Details}
\label{subsec:impl}

{\bf Precomputation of Motion Barcodes.} The motion barcodes were computed for points on the silhouette boundaries which are also on the convex hull boundary, called {\em candidate points}. 180 angles are sampled every $2\degree$, each angle defines a tangent line to the silhouette through one of the candidate points. This results in 180 candidate lines per frame, and a motion barcode is computed for all these lines. In a video having $N$ frames, each motion barcode is a binary sequence of length $N$. A barcode matrix is defined for each frame having 180 rows and $N$ columns. Each column represents a frame, and each row represents a tangent line. Each row is the motion barcode of the corresponding candidate line. Given corresponding frames of two cameras frames, the distance between all possible pairs of candidate lines is computed by multiplying their motion barcode matrices, resulting in an 180$\times$180 affinity matrix of candidate lines.

Given the $N$ pairs of frames of two cameras, we extract for each frame the single pair of candidate lines having the highest barcode correlation. This results in $N$ pairs of possible matching epipolar lines, each having higher barcode correlation.
 
{\bf RANSAC Sampling.}  The efficiency of fundamental matrix computation can be broken into the initialization cost, the number of hypotheses needed to be generated, the cost of generating an hypothesis and the cost of hypothesis verification. In both methods the cost of the model verification phase is identical as it is indifferent to the model generation.  The comparison is therefore the number of RANSAC hypotheses required by each method. In the following we provide a detailed description.

Generating the hypothesized model is as follows. In our method, three matching pairs having high barcode correlation were randomly selected from the pre-computed table of barcode correlations between all pairs of candidate lines. For the Sinha method, as described in \cite{sinha2010camera}, two matching hypothesized lines were extracted based on sampling the directions of the tangents in one frame. The third matching pair of lines was computed using the epipole generated by the first two lines, and a tangent to a silhouette in another frame. Given three proposals for corresponding epipolar lines, the fundamental matrix was computed using the method described in \cite{sinha2010camera}. 
The computation of the third matching pair of lines by the generated epipoles could be applied in our approach as well, requiring selection of only two matching lines instead of three. This could improve the accuracy of the method. On the other hand, it  requires additional computations for finding the exact tangents in each RANSAC iteration. We empirically saw that sampling three lines is faster than sampling two lines together with  the additional tangent computations.          

The cost of each RANSAC iteration depends on (a) lines match generation and (b) the computation of the fundamental matrix from the epipolar line homography and the epipoles. For the motion barcode method the first part is instantaneous as it involves only index selection, since matching pairs of lines are computed beforehand. For the baseline method each iteration introduces the computation of six tangents, where the computation of the last pair of tangents involves finding the frontier points with respect to the hypothesized epipoles.  The second part is  the same for all the methods and introduces the major cost of each iteration. We assume that the first part is instantaneous also in the baseline method and consider the cost of each iteration as the cost of the second part.    

Computing the motion barcode distance between all pairs of candidate lines adds computation efforts to our method. This computational cost was equivalent to 35 iterations of RANSAC, which we added to the cost of our method.
 
{\bf Ground Truth points.}  For accurate evaluation of the symmetric epipolar distance we extracted matching frontier points across different views, using the given ground-truth silhouettes and the given ground truth fundamental matrix.  For each frame, points whose tangent line is within an angular deviation of $1\degree$ of the true epipolar line were extracted. A pair of points was considered  frontier if their epipolar distance using the known fundamental matrix is less than $0.01$.  This results in a cloud of points that might be spread out unevenly. From these points we sampled ground truth points that have a distance of at least 15 pixels from each other, resulting in several dozen point, well spread out, per view. 

\section{Concluding Remarks}
\label{sec:conclusion}

Motion barcodes were introduced as efficient temporal signatures for lines, signatures which are viewpoint invariant for matching epipolar lines.
The effectiveness of motion barcodes was demonstrated in camera calibration using candidate epipolar lines. In this case, computing candidate fundamental matrices only from candidate lines that have matching motion barcodes, reduced computational costs by about two orders of magnitude.

\ifcvprfinal{
\pagestyle{empty}
~\\
\noindent
{\bf Acknowledgment.} This research was supported by Google, by Intel ICRI-CI, by DFG, and by the Israel Science Foundation.
}\fi

{\small
\bibliographystyle{ieee}
\bibliography{egbib}
}

\end{document}